\documentclass[10pt,twocolumn,letterpaper]{article}

\usepackage{cvpr}              
\usepackage{graphicx}
\usepackage{subcaption}
\usepackage{tabularx}
\usepackage{placeins}
\usepackage{inputenc}

\usepackage{placeins}
\usepackage{pifont}
\newcommand{\cmark}{\ding{51}} 
\newcommand{\xmark}{\ding{55}} 
\usepackage{booktabs}    
\usepackage{tabularx}    
\usepackage{multirow}
\usepackage{makecell}

\definecolor{cvprblue}{rgb}{0.21,0.49,0.74}
\usepackage[pagebackref,breaklinks,colorlinks,allcolors=cvprblue]{hyperref}

\def\paperID{*****} 
\def\confName{CVPR}
\def\confYear{2026}

\title{One Patch is All You Need: Joint Surface Material Reconstruction and Classification from Minimal Visual Cues}

\author{
Sindhuja Penchala,
Gavin Money,
Gabriel Marques,
Samuel Wood,
Jessica Kirschman, \\
Travis Atkison,
Shahram Rahimi,
Noorbakhsh Amiri Golilarz\\[2pt]
\normalsize{Department of Computer Science, The University of Alabama, Tuscaloosa, AL, USA}\\
}

\begin{document}
\maketitle
\begin{abstract}
Understanding material surfaces from sparse visual cues is critical for applications in robotics, simulation, and material perception. However, most existing methods rely on dense or full-scene observations, limiting their effectiveness in constrained or partial view environment. To address this challenge, we introduce SMARC, a unified model for Surface MAterial Reconstruction and Classification from minimal visual input. By giving only a single 10\% contiguous patch of the image, SMARC recognizes and reconstructs the full RGB surface while simultaneously classifying the material category. Our architecture combines a Partial Convolutional U-Net with a classification head, enabling both spatial inpainting and semantic understanding under extreme observation sparsity. We compared SMARC against five models including convolutional autoencoders \cite{CAE}, Vision Transformer (ViT) \cite{EnhancedVit}, Masked Autoencoder (MAE) \cite{Mae}, Swin Transformer \cite{Swin}, and DETR \cite{DETR} using Touch and Go dataset \cite{touchgo} of real-world surface textures. SMARC achieves state-of-the-art results with a PSNR of 17.55 dB and a material classification accuracy of 85.10\%. Our findings highlight the advantages of partial convolution in spatial reasoning under missing data and establish a strong foundation for minimal-vision surface understanding.
\end{abstract}
\section{Introduction}
\label{sec:intro}

Surface material reconstruction and categorization are essential for robotic perception and physical contact with the environment \cite{Advapp}. These abilities let robots to recognize textures, hardness, and reflectivity,important clues that influence how a robot perceives, manipulates, and responds to real-world surfaces \cite{Pix2vox}. Understanding whether a surface is metallic, soft, or granular has a direct impact on applied force, grip stability, and tool trajectory in robotic manipulation activities like grasping, drilling, and polishing \cite{Roboapplication}. However, visual inputs obtained in unstructured or congested surroundings are frequently insufficient due to occlusion, sensor noise, or limited views. Such limited observations make it difficult to derive precise material properties or reconstruct surface appearance from incomplete data. As a result, there is a growing need for perception frameworks that can reason beyond limited seen pixels and infer missing information in order to accomplish accurate surface understanding and reconstruction in robotic systems \cite{RobotSurfrec}.

Early image reconstruction methods relied heavily on convolutional architectures, which excel at modeling local spatial patterns and capturing fine-grained texture details. Classical frameworks such as Convolutional Autoencoders \cite{CAE} and U-Net established the foundation for dense image restoration tasks including denoising, inpainting, and semantic segmentation. These architectures progressively extract hierarchical features, enabling them to model localized geometric cues and edge continuity effectively. However, their dependence on local receptive fields restricts global reasoning, making them less effective when large portions of an image are missing or corrupted \cite{Deepcnn}.

The emergence of transformer-based vision models \cite{Dosovitskiy20} has significantly expanded the scope of image understanding. Vision Transformers (ViT) \cite{EnhancedVit} introduced self-attention mechanisms that learn long-range dependencies and global feature interactions, while hierarchical extensions such as the Swin Transformer \cite{Swinir} enhanced scalability by integrating multi-resolution windowed attention. Similarly, Masked Autoencoders (MAE) \cite{Mae} advanced self-supervised representation learning by reconstructing missing patches from visible regions, leading to strong generalization across downstream tasks. DETR \cite{DETR} further applied the transformer paradigm to object recognition and localization, reformulating perception as a set-based prediction problem through end-to-end attention modeling. Collectively, these advances have demonstrated powerful performance in complete-view settings, where sufficient visual context is available for learning structure and semantics jointly.

Despite their success, these architectures remain limited in scenarios where the input is highly sparse or incomplete. Convolution based models tend to produce over-smoothed and textureless output because local kernels cannot infer long-range context beyond visible regions\cite{Scnn}. Transformer based models, though effective at global reasoning, rely on dense token embeddings and struggle when visual evidence is insufficient to establish spatial correspondence \cite{Dosovitskiy20}. The random masking strategy employed by MAE disrupts spatial continuity, resulting in coarse reconstructions that lack fine structural fidelity. Consequently, existing frameworks are not well-suited for perception tasks that require robust reasoning under heavy occlusion such as surface material reconstruction and classification in robotic environments, where accurate inference from minimal visual cues is essential.

To overcome these limitations, we developed SMARC, a dual purpose framework that integrates partial convolutions and mask propagation within a U-Net-like architecture. Unlike prior methods, SMARC operates exclusively on valid, continuous (unmasked) pixels and dynamically updates the mask at each layer to maintain spatial consistency. This design enables the network to reconstruct missing regions with high structural fidelity while performing joint surface classification. Through this mask aware learning strategy, SMARC bridges the gap between incomplete perception and reliable material understanding. The key contributions of this work are as follows:

\begin{itemize}
\item \textbf{SMARC} is proposed as a unified framework for joint surface reconstruction and classification from only 10\% visible input.
\item A mask aware Partial Convolutional U-Net is designed to dynamically update visibility masks, guiding both reconstruction and classification.
\item The effectiveness of SMARC is demonstrated on the Touch and Go dataset, achieving 17.55\% PSNR and 85.10\% accuracy, outperforming five state of the art models.
\item \textbf{SMARC} exhibits superior computational efficiency, achieving the fastest inference speed of 19.1M parameters per second, demonstrating its scalability and real time processing capability in robotics.
\end{itemize}

The rest of paper is structured as follows. 
Section 2 introduces the proposed SMARC architecture, detailing the partial convolutional U-Net backbone, classification head, and training strategy. Section 3 presents experimental results on the Touch and Go dataset, comparing reconstruction and classification metrics. Finally, Section 4 concludes the paper and discusses potential directions for extending minimal view surface understanding in robotics and vision applications.

\section{Methodology}
\label{sec:methodology}

\begin{figure*}
  \centering
  \includegraphics[width=\linewidth,height=1.0\textheight,keepaspectratio]{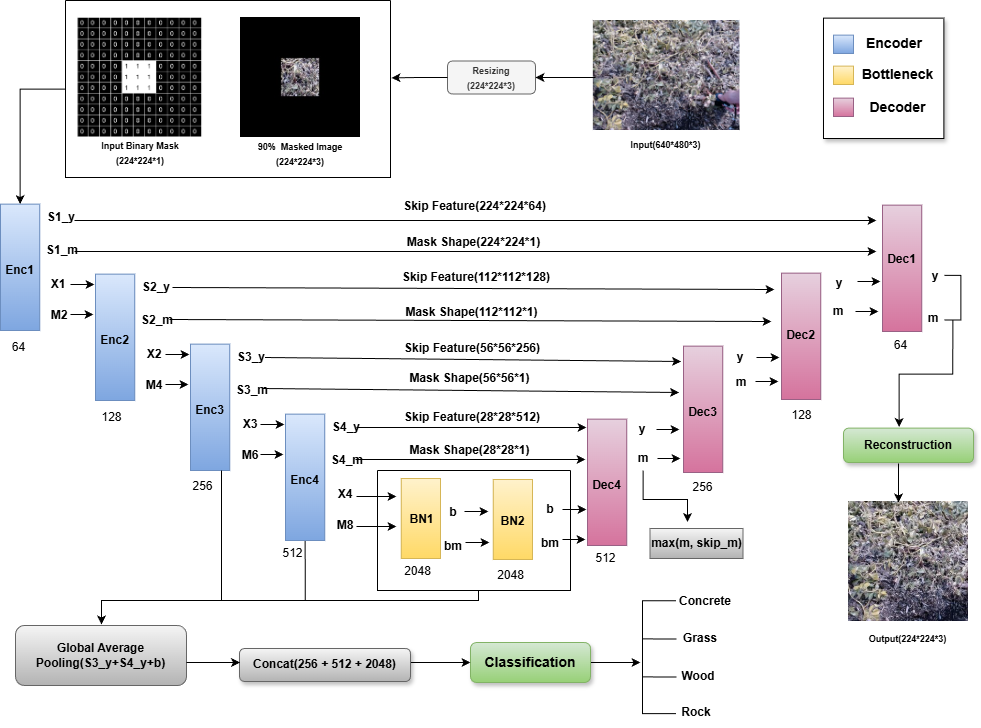}
  \caption{Overview of \textbf{SMARC}. The network follows an encoder–bottleneck–decoder design with partial convolutions and explicit mask propagation. Skip connections fuse encoder features into the decoder for reconstruction, while a multi-scale head pools features from $s_3^y$, $s_4^y$, and the bottleneck $b$ for surface classification. The model restores occluded regions in the RGB image and simultaneously predicts the material class.}
  \label{fig:smarc_overview}
\end{figure*}

We propose SMARC, a dual-purpose architecture built upon a modified U-Net framework with partial convolutions to jointly perform image inpainting and surface classification. The model is designed to restore images with missing or occluded regions while simultaneously identifying the underlying surface type such as concrete, grass, wood, or rock. SMARC follows an encoder–bottleneck–decoder structure enhanced with skip connections, a dedicated mask propagation path, and a multi-scale classification head (shown in  Figure \ref{fig:smarc_overview}). Within this design, partial convolutions ensure that feature extraction and reconstruction operate exclusively on valid pixels, thereby minimizing artifacts and preserving structural integrity in masked areas. The decoder progressively restores the complete RGB image, while the classification branch fuses multi level semantic features to enable accurate and relevant surface recognition.

\subsection{Encoder Architecture}

The encoder consists of four hierarchically arranged blocks, each designed to extract semantically rich features from partially masked input images. At each level, the encoder receives both a feature map and a corresponding binary mask as input that identify valid (unmasked) regions. This mask is propagated and refined in parallel with the feature map, enabling robust learning from incomplete observations. Each block operates using \textit{partial convolutional layers} (PConv) and incorporates channel wise attention via \textit{Squeeze-and-Excitation (SE)} modules.

Let the input image be \( x \in \mathbb{R}^{H \times W \times 3} \), and the respective binary mask be \( m \in \{0,1\}^{H \times W \times 1} \). The encoder processes these through four sequential blocks \(\{\text{Enc}_1, \text{Enc}_2, \text{Enc}_3, \text{Enc}_4\}\), each consisting of two partial convolution layers with kernel size \(3 \times 3\), followed by ReLU activations and an SE attention module.

Each encoder block produces(referenced in table \ref{tab:variables}):
\begin{itemize}
    \item \( s_i^y \): the skip connection feature map,
    \item \( s_i^m \): the updated binary mask, and
    \item \( x_i, m_i \): the spatially downsampled features and masks for the next block.
\end{itemize}

The number of output channels increases with depth, starting with 64, 128, 256, and 512, respectively. Feature maps are downsampled using average pooling, while masks are downsampled using max pooling to preserve binary validity. By processing both features and their corresponding masks together, the encoder learns rich, multi-scale representations that remain spatially consistent and aware of missing regions. These representations provide a strong foundation for the decoder and classification branches, enabling accurate image reconstruction and reliable scene understanding.


\subsection{Bottleneck Design}

The bottleneck component operates at the model's most reduced spatial resolution and acts as a bridge between the encoder and decoder stages. Its goal is to refine encoded features and provide global context that can help in accurate reconstruction and classification. The bottleneck gets its input from the deepest feature map and the binary mask output from the last encoder block, denoted \( x_4 \in \mathbb{R}^{14 \times 14 \times 512} \) and \( m_4 \in \{0,1\}^{14 \times 14 \times 1} \), respectively. These are passed through two bottlenecks, each bottleneck block applies two partial convolution layers with increasing dilation rates. The first block uses a dilation rate of 2, and the second uses a rate of 4. This is followed by ReLU activations and a squeeze-and-excitation (SE) module for adaptive channel wise attention. The use of partial convolutions ensures that all computations are based on valid (unmasked) pixels, with the binary mask being updated accordingly after each layer.

This design allows the network to increase its receptive field without further downsampling, allowing it to aggregate spatial information across a larger context. It is an essential property for tasks involving incomplete or occluded inputs. The output of the bottleneck stage includes:
\begin{itemize}
    \item A feature tensor \( b \in \mathbb{R}^{14 \times 14 \times 2048} \), which encodes high-level semantic information.
    \item An updated binary mask \( b_m \in \{0,1\}^{14 \times 14 \times 1} \), which indicates the validity of each spatial location after context aggregation.
\end{itemize}

These outputs are passed forward to both the decoder for image reconstruction and the classification head for scene recognition. By maintaining both the feature and mask paths, the bottleneck preserves structural knowledge and makes sure that the next layers work only on important areas.

\begin{table}[t]
\centering
\caption{Key intermediate variables in the SMARC architecture. Shapes are height~$\times$~width~$\times$~channels.}
\label{tab:variables}
\renewcommand{\arraystretch}{1.1}
\setlength{\tabcolsep}{4pt}
\scriptsize
\begin{tabularx}{\linewidth}{@{}l c X@{}}
\toprule
\textbf{Variables} & \textbf{Shape} & \textbf{Description} \\
\midrule
\begin{tabular}[c]{@{}l@{}}\texttt{S1\_y, S2\_y,}\\ \texttt{S3\_y, S4\_y}\end{tabular}
& $224{\times}224{\times}64 \rightarrow 28{\times}28{\times}512$
& Skip features from the four encoder blocks; provide multi-scale context to the decoder and the classification head. \\
\addlinespace[2pt]
\begin{tabular}[c]{@{}l@{}}\texttt{X1, X2,} \\ \texttt{X3, X4}\end{tabular}
& $112{\times}112{\times}64 \rightarrow 14{\times}14{\times}512$
& Downsampled encoder features (post-pooling) passed forward to the next encoder stage, with \texttt{X4} feeding the bottleneck. \\
\addlinespace[2pt]
\begin{tabular}[c]{@{}l@{}}\texttt{S1\_m, S2\_m,}\\ \texttt{S3\_m, S4\_m}\end{tabular}
& $224{\times}224{\times}1 \rightarrow 28{\times}28{\times}1$
& Progressive binary masks propagated alongside skip features; preserve spatial validity for partial convolutions. \\
\addlinespace[2pt]
\begin{tabular}[c]{@{}l@{}}\texttt{M1, M2,} \\ \texttt{M3, M4}\end{tabular}
& $112{\times}112{\times}1 \rightarrow 14{\times}14{\times}1$
& Downsampled masks (post-pooling) aligned with \texttt{X1 to X4}; \texttt{M4} is passed to the bottleneck. \\
\addlinespace[2pt]
\texttt{b}, \texttt{bm}
& $14{\times}14{\times}2048$, $14{\times}14{\times}1$
& Bottleneck feature and mask after dilated PConv blocks; provide global semantic context for decoder and classifier. \\
\addlinespace[2pt]
\texttt{y}, \texttt{m}
& $224{\times}224{\times}64$, $224{\times}224{\times}1$
& Final decoder feature and propagated mask before RGB output; mask merged via element-wise max during upsampling. \\
\bottomrule
\end{tabularx}
\end{table}

\subsection{Decoder Architecture}

Thereafter, the decoder takes the high-level feature tensor \( b \in \mathbb{R}^{14 \times 14 \times 2048} \) and its corresponding binary mask \( b_m \in \{0,1\}^{14 \times 14 \times 1} \) produced by the bottleneck. Its role is to progressively recover spatial details and reconstruct the complete image. Each decoding stage doubles the spatial resolution while reducing the number of channels, following a symmetric structure with the encoder. In each stage, the feature map is first upsampled using a transposed convolution with a stride of 2. The upsampled feature is then concatenated with the corresponding skip feature from the encoder. This allows the model to reintroduce fine grained details that were captured earlier. The associated masks are merged using an element-wise maximum operation to ensure that any valid region from either path remains active. The combined features and masks are refined using a \texttt{pconv\_block}, which applies two partial convolutions with ReLU activations and a squeeze-and-excitation module for channel recalibration.

This process is repeated over four stages (\texttt{dec4}–\texttt{dec1}) with 512, 256, 128, and 64 filters respectively, restoring the spatial size from \(14 \times 14\) to \(224 \times 224\). Finally, a \(1 \times 1\) convolution followed by a sigmoid activation produces the reconstructed RGB image \(\hat{I} \in \mathbb{R}^{224 \times 224 \times 3}\), with pixel intensities normalized to the range \([0,1]\). By integrating semantic context from the bottleneck with local spatial cues from the encoder, the decoder produces natural and structurally consistent reconstructions, even in areas that were initially masked or missing.

\subsection{Multi Scale Classification Head}

In addition to image reconstruction, the network includes a classification branch that predicts the category of the input image using multi scale feature representations. This branch supplements the decoder by utilizing hierarchical information from different encoder stages. Let the feature maps from the third and fourth encoder blocks and the bottleneck be denoted as 
\( s_3^y \in \mathbb{R}^{56 \times 56 \times 256} \),
\( s_4^y \in \mathbb{R}^{28 \times 28 \times 512} \), and 
\( b \in \mathbb{R}^{14 \times 14 \times 2048} \), respectively.
These multi scale characteristics from mid-level textures to high-level semantic contexts. Each of these feature maps is processed through a \textit{Global Average Pooling} (GAP) layer to convert the spatial maps into compact feature vectors. The resulting vectors are concatenated to form a unified representation as follows:

\begin{equation}
f_{\text{cls}} = \text{Concat}\big(\text{GAP}(s_3^y),\, \text{GAP}(s_4^y),\, \text{GAP}(b)\big)
\label{eq:classification_head}
\end{equation}

The combined feature vector is passed through fully connected layers with ReLU activations and dropout regularization to improve generalization. The final dense layer employs a softmax activation to output the class probabilities corresponding to the target categories. Each classification head component consists as follows:
\begin{itemize}
    \item Inputs: multi-scale encoder features \((s_3^y, s_4^y, b)\).
    \item Processing: global pooling, concatenation, and fully connected layers.
    \item Output: softmax-based class probabilities for surface material prediction.
\end{itemize}

By integrating input across various dimensions, the classification head makes use of detailed spatial cues, improving the model's ability to execute reconstruction and classification simultaneously without adding computational burden.





\section{Experimental Analysis}
\label{sec:expsetup}

To evaluate the effectiveness of the proposed SMARC model, we conduct a comparative analysis consistent with standard practices in sparse input vision research. In this evaluation, we consider five representative architectures including Convolutional Autoencoder \cite{CAE}, ViT \cite{EnhancedVit}, Masked Autoencoder \cite{Mae}, Swin \cite{Swinir}, and DETR \cite{DETR}. They serve as strong baselines for either reconstruction or recognition tasks. However, these models are not inherently designed to perform both objectives under extreme input sparsity. By comparing their performance under identical minimal view conditions, the study demonstrates the robustness and adaptability of SMARC in reconstructing and classifying. The following sections outline the data preparation, training methodology, and evaluation metrics employed to ensure consistency and fairness across all comparative experiments.

\subsection{Experimental Setup}

\subsubsection{Data  Preparation}

Our training data derives from the Touch and Go dataset, which contains video examples of various textile and surface materials captured in natural environments~\cite{touchgo}. Each video was decomposed into individual frames, which were then resized to $224 \times 224$ RGB images. These images were processed through a masking script that preserved only the central 10\% of pixels, simulating highly occluded conditions. The binary mask was passed in parallel with the input image and dynamically updated at each layer to track valid spatial regions, enabling the use of partial convolutions that operate exclusively on visible pixels. The dataset was partitioned into 60\% (1753 images) for training, 20\% (584 images) for validation, and 20\%  (584 images) for testing, ensuring balanced representation of all material categories. This configuration enhances the model's resilience to incomplete observations and supports consistent evaluation.


Each input sample consists of two components as follows:
\begin{itemize}
    \item A masked RGB image of size 224×224×3, and
    \item A corresponding binary mask of shape 224×224×1,  indicates which pixels are valid (1) and which are missing (0).
\end{itemize}

\subsubsection{Training Strategy}




The network was trained in two sequential phases to ensure stable optimization and balanced learning across the encoder, bottleneck, decoder, and classification head.

\textbf{Phase A – Head Warm-Up.}  
In the first stage, only the classification head was trained while all encoder, decoder, bottleneck, and reconstruction layers were frozen. This step allowed the classifier to adapt to high-level semantic representations without disrupting the pretrained feature extraction layers. Training was performed using the Adam optimizer with a learning rate of $2\times10^{-4}$ for 10~epochs. To enhance robustness, class weights were computed from label frequencies to counter data imbalance, and data augmentation (random flips, rotations, and brightness/contrast perturbations) was applied to improve generalization under masked conditions.

\textbf{Phase B – Fine-Tuning.}  
All layers were subsequently unfrozen and fine-tuned end-to-end with a reduced learning rate of $1\times10^{-4}$. This phase jointly optimized reconstruction and classification objectives. The total loss combined a pixel-wise Mean Absolute Error (MAE) for local fidelity, a perceptual loss for high-level texture consistency, and a categorical cross-entropy term for scene classification.

During training, mild overfitting was observed as training loss decreased faster than validation accuracy. This indicated that the network was memorizing training patterns rather than generalizing effectively to unseen data. To address this issue and improve robustness, several complementary strategies were introduced throughout the training process:


\begin{itemize}
    \item \textbf{Two-Phase Optimization.} A head warm-up stage trains only the classifier for 10 epochs at $\text{LR}=2\times10^{-4}$ (encoder, bottleneck, decoder, and RGB head frozen), followed by full end-to-end fine-tuning at $\text{LR}=1\times10^{-4}$ for up to 150 epochs (Adam; seed $=42$; batch size $=16$).
    \item \textbf{Data Augmentation.} Rotations by $90^\circ$ increments ($k\in\{0,1,2,3\}$), horizontal/vertical flips, mild color jitter (brightness $\pm 0.06$, contrast $[0.90,1.10]$, saturation $[0.90,1.10]$), and light Gaussian noise ($\sigma\in[0,0.02]$) are applied within the \texttt{tf.data} pipeline. These transforms increase effective sample diversity per epoch while preserving mask alignment.
    \item \textbf{Imbalance Handling.} Inverse-frequency \emph{class weights} derived from the training split are applied to the classification loss to prevent bias toward majority classes.
    \item \textbf{Regularization.} L2 weight decay with coefficient $10^{-4}$ is applied to convolutional, transposed-convolutional, and dense kernels. The classification head uses dropout with rate $p=0.25$. Batch normalization is used throughout to stabilize activations and gradients.
    \item \textbf{Early Stopping and LR Scheduling.} Training employs early stopping on validation accuracy with patience $=18$ epochs (best weights restored). A ReduceLROnPlateau scheduler halves the learning rate after 8 stagnant epochs (factor $0.5$, minimum $\text{LR}=10^{-6}$).
    \item \textbf{Multi-Task Loss.} The total objective combines a mask-weighted MAE for reconstruction (weight $\lambda_{\text{rgb}}=0.25$) with categorical cross-entropy for classification using label smoothing $\varepsilon=0.05$, plus L2 regularization from layer kernels.
\end{itemize}

These strategies collectively improved generalization, reduced variance between training and validation performance, and ensured stable convergence across both reconstruction and classification tasks.

\begin{table}[t]
\centering
\caption{Reconstruction performance of all models on the Touch and Go \cite{touchgo} dataset using four metrics. ↑ indicates higher is better, ↓ indicates lower is better.}
\label{tab:reconstruction}
\begin{tabular}{lcccc}
\toprule
\textbf{Model} & \textbf{PSNR ↑} & \textbf{SSIM ↑} & \textbf{MSE ↓} & \textbf{MAE ↓} \\
\midrule
Conv AE \cite{CAE}   & 15.75          & 0.5458          & 0.0330          & 0.1382          \\
MAE \cite{Mae}            & 17.01          & 0.5265          & 0.0267          & 0.1132          \\
ViT \cite{EnhancedVit}               & 16.00          & 0.5211          & 0.0314          & 0.1342          \\
Swin      \cite{Swinir}         & 16.39          & 0.5334          & 0.0278          & 0.1221          \\
DETR   \cite{DETR}            & 17.07          & 0.5289          & 0.0252          & 0.1115          \\
\textbf{SMARC} & \textbf{17.55} & \textbf{0.5733} & \textbf{0.0223} & \textbf{0.0987} \\
\bottomrule
\end{tabular}
\end{table}

\begin{table}[t]
\centering
\caption{Classification performance comparing all models. $\uparrow$ indicates higher is better.}
\label{tab:classification}
\setlength{\tabcolsep}{5pt}
\renewcommand{\arraystretch}{1.1}
\small
\resizebox{\linewidth}{!}{%
\begin{tabular}{@{}lcccc@{}}
\toprule
\textbf{Model} & \textbf{Accuracy $\uparrow$} & \textbf{Precision $\uparrow$} & \textbf{Recall $\uparrow$} & \textbf{F1-Score $\uparrow$} \\
\midrule
Conv AE   & 0.8373 & 0.8371 & 0.8373 & 0.8362 \\
MAE       & 0.7466 & 0.7560 & 0.7466 & 0.7465 \\
ViT       & 0.7432 & 0.7503 & 0.7432 & 0.7443 \\
Swin      & 0.7192 & 0.7366 & 0.7192 & 0.7232 \\
DETR      & 0.7243 & 0.7351 & 0.7243 & 0.7253 \\
\textbf{SMARC} & \textbf{0.8510} & \textbf{0.8568} & \textbf{0.8510} & \textbf{0.8514} \\
\bottomrule
\end{tabular}%
}
\end{table}

\begin{table}[t]
\centering
\caption{Inference time calculated on test dataset. Parameters-per-second (M/s) denotes millions of parameters processed per second.}
\label{tab:runtime}
\setlength{\tabcolsep}{5pt}
\renewcommand{\arraystretch}{1.1}
\small
\resizebox{\linewidth}{!}{%
\begin{tabular}{@{}lcccc@{}}
\toprule
\textbf{Model} &
\textbf{\makecell{Params\\(M)}} &
\textbf{\makecell{Inference.\\Time (s/img)}} &
\textbf{\makecell{Total\\Time (s)}} &
\textbf{\makecell{Params/sec\\(M/s)}} \\
\midrule
Conv AE   & 7.61   & 0.0051 & 2.98 & 2.53 \\
MAE       & 35.28  & 0.0069 & 4.03 & 8.75 \\
ViT       & 29.19  & 0.0051 & 2.98 & 9.68 \\
Swin      & 1.50   & 0.0114 & 6.66 & 0.23 \\
DETR      & 7.87   & 0.0052 & 3.04 & 2.59 \\
\textbf{SMARC} & \textbf{145.07} & 0.0130 & 7.59 & \textbf{19.10} \\
\bottomrule
\end{tabular}%
}
\end{table}



\begin{figure*}[t]
\centering
\begin{subfigure}[t]{0.32\textwidth}
    \includegraphics[width=\linewidth]{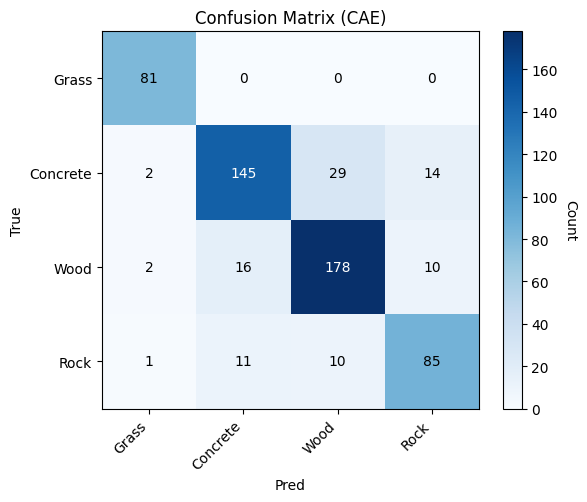}
    \caption{Conv Autoencoder}
\end{subfigure}
\hfill
\begin{subfigure}[t]{0.32\textwidth}
    \includegraphics[width=\linewidth]{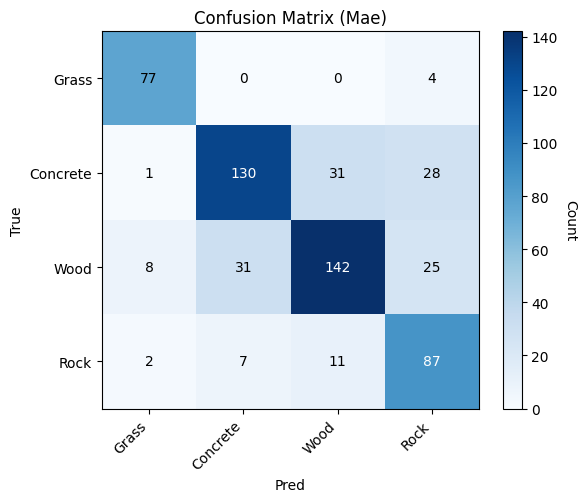}
    \caption{MAE}
\end{subfigure}
\hfill
\begin{subfigure}[t]{0.32\textwidth}
    \includegraphics[width=\linewidth]{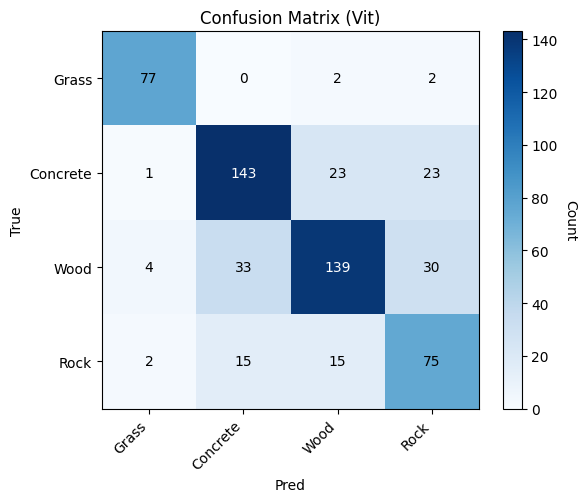}
    \caption{ViT}
\end{subfigure}

\vspace{1em}

\begin{subfigure}[t]{0.32\textwidth}
    \includegraphics[width=\linewidth]{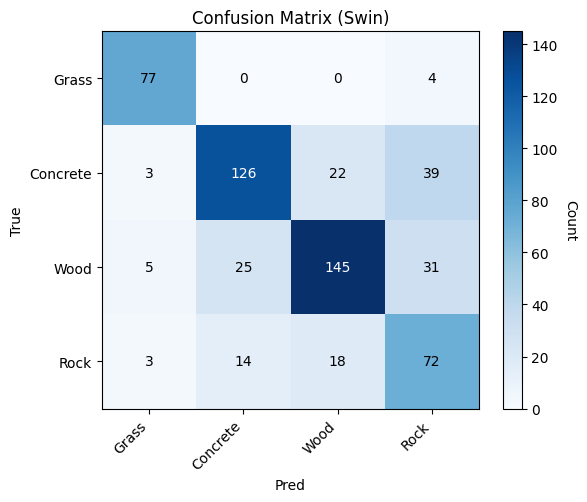}
    \caption{Swin}
\end{subfigure}
\hfill
\begin{subfigure}[t]{0.32\textwidth}
    \includegraphics[width=\linewidth]{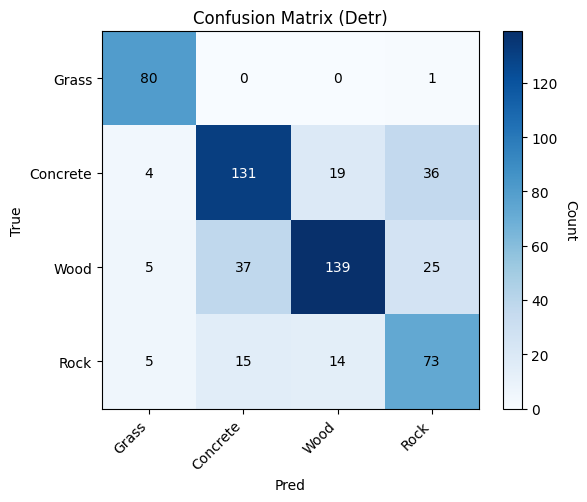}
    \caption{DETR}
\end{subfigure}
\hfill
\begin{subfigure}[t]{0.32\textwidth}
    \includegraphics[width=\linewidth]{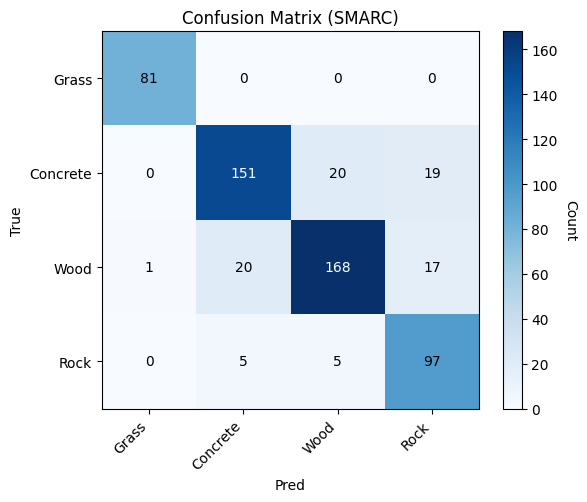}
    \caption{SMARC}
\end{subfigure}

\caption{Confusion matrices of all six models evaluated on the Touch and Go dataset using 10\% input crops. SMARC (f) shows strong diagonal dominance, reflecting robust class-wise performance.}
\label{fig:confusion_matrices}
\end{figure*}

\subsubsection{Metrics}
To quantitatively assess the performance of SMARC, we used a set of metrics that capture both reconstruction fidelity and classification accuracy. Following the training setup, reconstruction quality is evaluated using Peak Signal-to-Noise Ratio (PSNR), Structural Similarity Index Measure (SSIM), Mean Absolute Error (MAE), and Mean Squared Error (MSE), each reflecting a different aspect of pixel and structural level restoration. For the classification branch, we evaluated accuracy, precision, recall and F1-score, averaged across all classes to account for label imbalance. These metrics together provide a balanced view of the model’s ability to both reconstruct missing regions and correctly identify surface materials. All results are computed on the held out test set using the best checkpoint obtained during training.

\begin{figure*}[t]
\centering
\begin{subfigure}[t]{0.32\textwidth}
    \includegraphics[width=\linewidth]{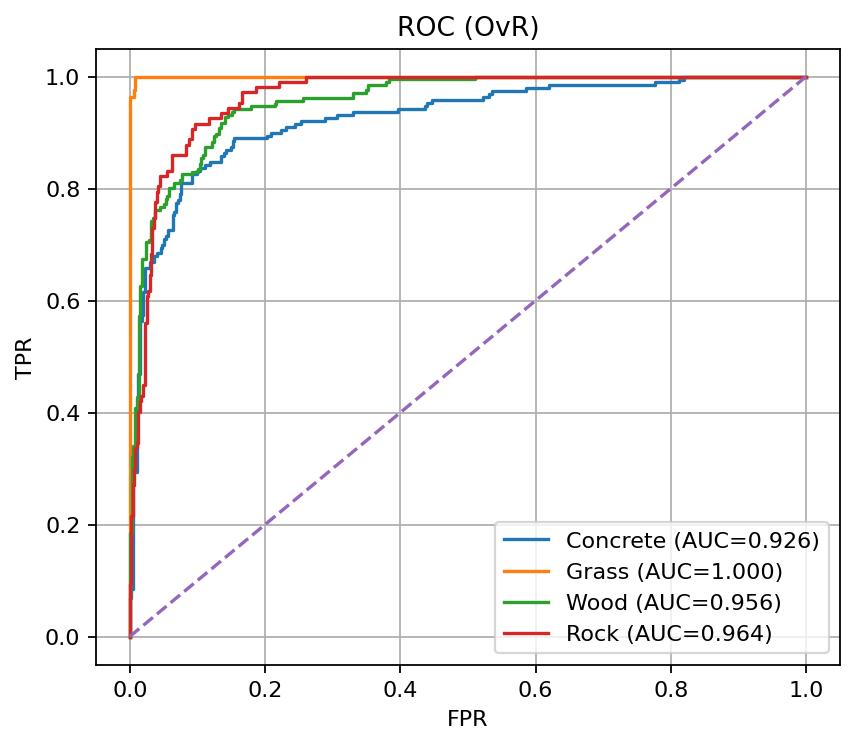}
    \caption{Conv Autoencoder}
\end{subfigure}
\hfill
\begin{subfigure}[t]{0.32\textwidth}
    \includegraphics[width=\linewidth]{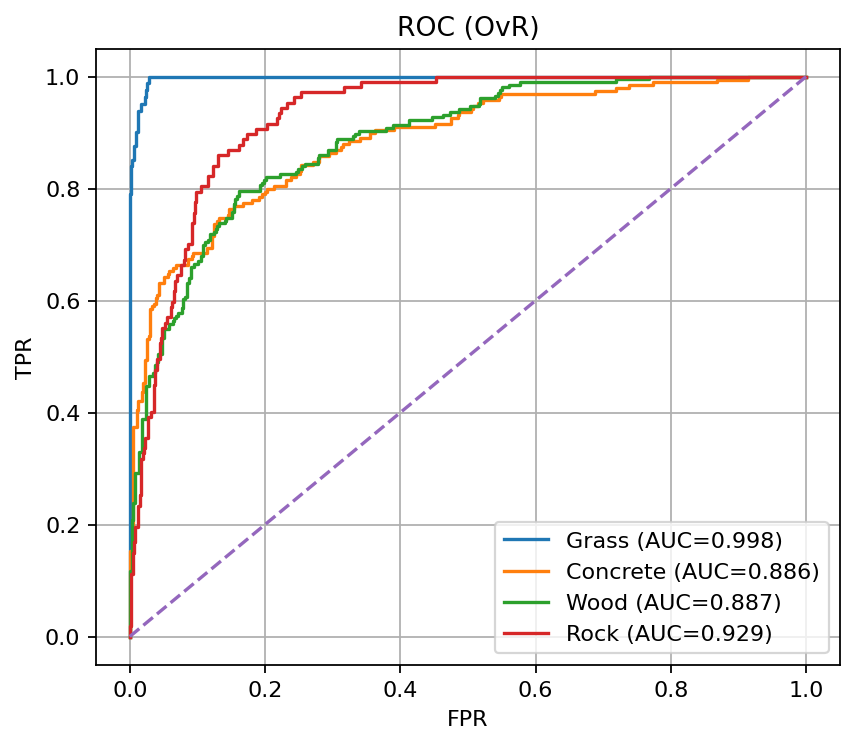}
    \caption{MAE}
\end{subfigure}
\hfill
\begin{subfigure}[t]{0.32\textwidth}
    \includegraphics[width=\linewidth]{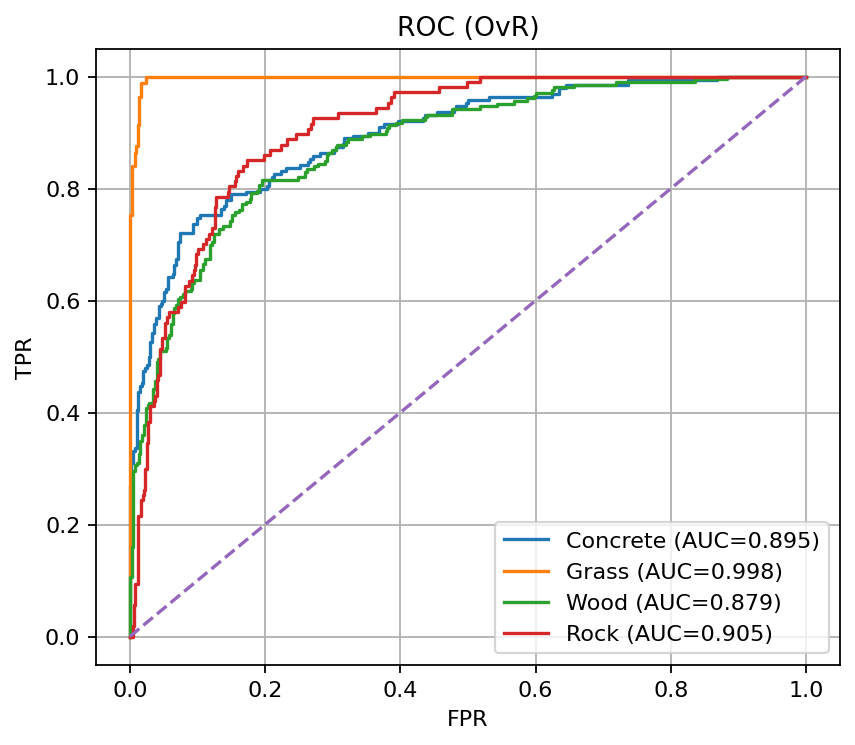}
    \caption{ViT}
\end{subfigure}

\vspace{1em}

\begin{subfigure}[t]{0.32\textwidth}
    \includegraphics[width=\linewidth]{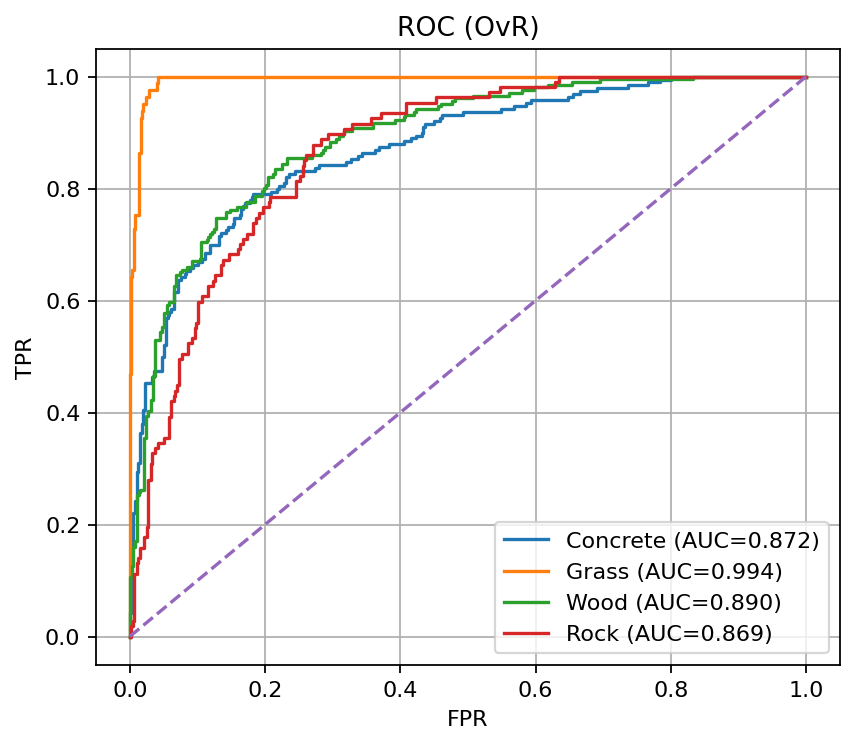}
    \caption{Swin}
\end{subfigure}
\hfill
\begin{subfigure}[t]{0.32\textwidth}
    \includegraphics[width=\linewidth]{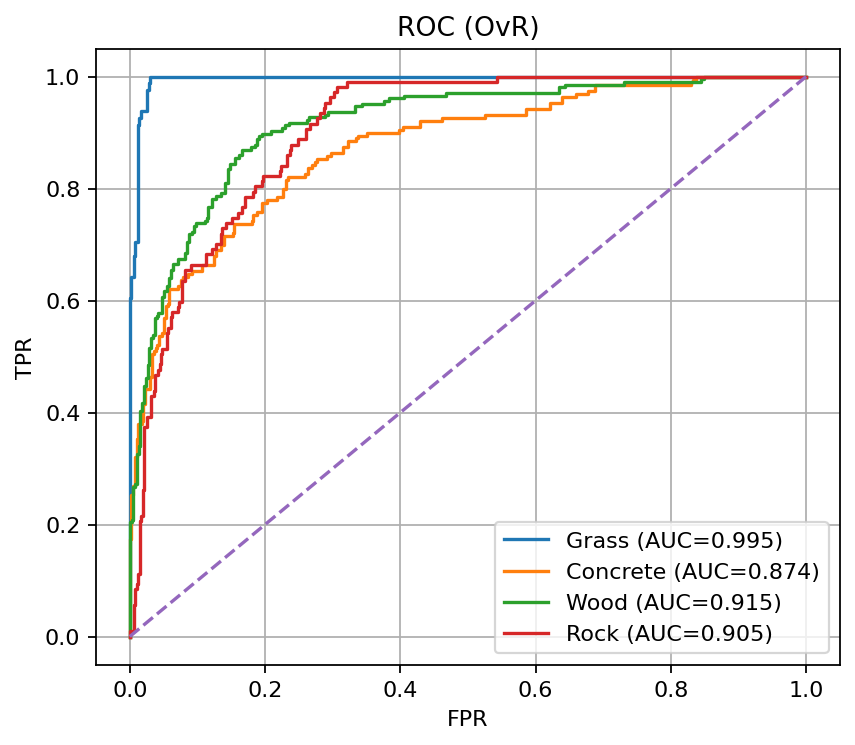}
    \caption{DETR}
\end{subfigure}
\hfill
\begin{subfigure}[t]{0.32\textwidth}
    \includegraphics[width=\linewidth]{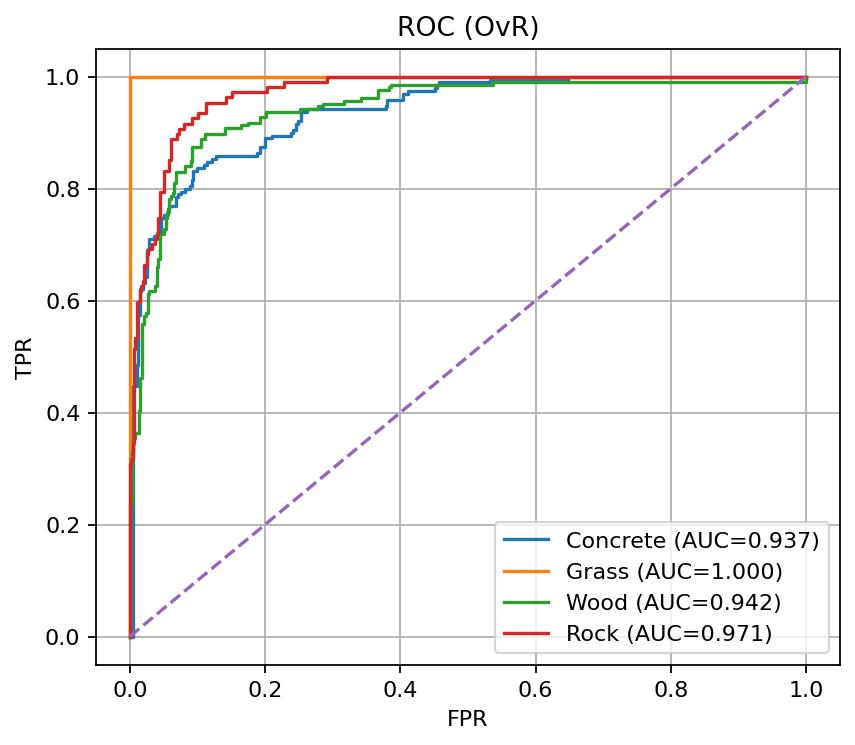}
    \caption{SMARC}
\end{subfigure}

\caption{ROC curves for all six models evaluated on the Touch and Go dataset. SMARC exhibits consistently higher AUC values across classes, highlighting its robust discrimination capabilities compared to other models.}
\label{fig:roc_curves}
\end{figure*}

\subsection{Results and Discussions}

In this section, We compare the performance of SMARC to five baseline architectures: Convolutional Autoencoder, Vision Transformer (ViT), Masked Autoencoder (MAE), Swin Transformer, and DETR. Our evaluation is based on three major criteria: reconstruction similarity, material classification performance, and model efficiency. All models are trained and evaluated on the Touch and Go dataset with the same minimal view configuration to ensure fair comparison.

\subsubsection{Reconstruction Quality}
To evaluate each model's performance to recover surface appearance from minimal visual input, we present four reconstruction metrics: PSNR, SSIM, MSE, and MAE. As seen in Table \ref{tab:reconstruction}, SMARC has the highest PSNR (17.55 dB) and SSIM values while preserving the lowest reconstruction errors in MSE and MAE. The gains over transformer-based models and autoencoders highlight the superiority of partial convolutions in reasoning over masked spatial inputs. Notably, transformer models such as MAE and ViT didn't perform well in this situation because they rely on distributed visual context, which is limited in our input regime.

\subsubsection{Classification Performance}
Table \ref{tab:classification} presents the classification performance of all models, evaluated using four standard metrics: accuracy, precision, recall, and F1-score. SMARC achieves the highest scores across all metrics, including an overall accuracy of 85.10\%, reflecting its ability to retain semantic understanding under severe visual occlusion. While transformer-based models such as ViT and Swin demonstrate competitive classification results, their performance in reconstruction tasks is noticeably weaker, underscoring the benefit of SMARC’s joint optimization for both appearance and material recognition. To further examine class level behavior, Figure 3 visualizes the confusion matrices for all models. SMARC exhibits the strongest diagonal alignment and the fewest off-diagonal errors, indicating more consistent and accurate material predictions across categories.

\begin{table*}[t]
\centering
\caption{Qualitative comparison of existing models and the proposed approach. Our method supports joint reconstruction and classification under low-visibility inputs using Touch and Go dataset \cite{touchgo}.}
\label{tab:qual_comp}
\setlength{\tabcolsep}{2pt}
\renewcommand{\arraystretch}{1.05}
\scriptsize
\begin{tabularx}{\linewidth}{@{}%
  >{\raggedright\arraybackslash}p{0.26\linewidth}%
  *{6}{>{\centering\arraybackslash}X}%
@{}}
\toprule
\textbf{Features} &
\textbf{MAE \cite{Mae}} &
\textbf{Surformer V1 \cite{SurformerV1}} &
\textbf{Uformer \cite{Uformer}} &
\textbf{Enhanced ViT \cite{EnhancedVit}} &
\textbf{PConv U\,-Net \cite{PConvUNet24}} &
\textbf{SMARC (Ours)} \\
\midrule
Multitasking (reconstruct + classify) & \xmark & \xmark & \xmark & \xmark & \xmark & \cmark \\
Vision-based (RGB)                    & \cmark & \cmark & \cmark & \cmark & \cmark & \cmark \\
Semantic labeling (compatible)        & \cmark & \cmark & \xmark & \xmark & \xmark & \cmark \\
Masked autoencoding                   & \cmark & \xmark & \xmark & \xmark & \xmark & \cmark \\
Explicit reconstruction               & \xmark & \cmark & \cmark & \cmark & \cmark & \cmark \\
\addlinespace[2pt]
How are they masked?                  & random patches & no mask & no mask & rows blacked-out / noise & irregular (user mask) & continuous patch \\
Input percentage (visible)            & 25\% & 100\% & 100\% & variable (task) & variable (mask) & 10\% \\
Output dimensionality                 & 2D image & class label & 2D image & 2D image & 2D image & 2D image + class label \\
\bottomrule
\end{tabularx}
\end{table*}

\subsubsection{Model Complexity and Inference Time}
Table \ref{tab:runtime} highlights the model complexity and inference time. SMARC achieves the highest performance but comes with the largest parameter count and a longer inference time of 13 ms. In contrast, models like Swin and Conv Autoencoder are significantly more lightweight, with ViT and Conv Autoencoder achieving the fastest inference speed at 5.1 ms. These results underscore the trade off between accuracy and computational efficiency, which must be considered based on deployment constraints. Among all models, the SMARC model achieved the fastest overall processing rate, handling approximately 19.1 million parameters per second, outperforming all baseline models in computational speed and efficiency. This faster inference rate is advantageous for real-time robotic applications, where rapid surface perception and material classification are critical\cite{RobotSurfrec}.

Additionally, to better understand each model's classification behavior beyond aggregate metrics, we examine them using confusion matrices and ROC curves. These diagnostic tools offer insight into the performance of discrimination per class and the consistency of prediction with minimal input.

\subsubsection{Confusion Matrix}

Figure~\ref{fig:confusion_matrices} illustrates the confusion matrices for all six models, offering a class-wise comparison of their classification behavior. Among all models, SMARC \ref{fig:confusion_matrices}(f) shows consistently strong class performance, particularly for classes like Grass and Rock, which are correctly classified with the lowest confusion. The Grass category is recognized with perfect accuracy, showing no off-diagonal errors. Although Concrete and Wood exhibit some degree of misclassification, Concrete occasionally predicted as Wood or Rock. This indicates a high level of confidence and robustness in SMARC’s predictions, even under highly occluded input. Compared to other models, its confusion matrix reflects balanced accuracy across all surface types and reduced ambiguity in decision boundaries, underscoring the strength of its joint reconstruction-classification design.

\subsubsection{ROC Curves}

Figure~\ref{fig:roc_curves} shows the ROC curves for all six models evaluated on the Touch and Go dataset. SMARC (f) consistently achieves superior AUC scores across all material categories, with perfect classification for Grass (AUC = 1.000) and high separability for Rock (AUC = 0.971), Concrete (AUC = 0.93) and Wood (AUC = 0.942). This indicates strong generalization and robust discrimination under limited visual input. In contrast, models such as Swin (d) and ViT (c) exhibit reduced performance, particularly for the Rock and Wood categories, with AUC values dropping below 0.90. Despite moderate performance by Conv Autoencoder and MAE, their curves reflect less consistent margins across classes. SMARC’s curves, by comparison, are tighter and steeper, further confirming its classification confidence and reliability in recognizing materials from sparse visual cues.

\subsubsection{Qualitative Comparison}

Table \ref{tab:qual_comp} contrasts SMARC with prior reconstruction and vision-based methods. While MAE, Uformer, Enhanced ViT, and PConv U-Net focus solely on pixel level restoration or denoising, they lack multitask capability and semantic coupling. Surformer V1 performs classification but without reconstruction or masking. In contrast, SMARC uniquely integrates continuous patch masking with a joint reconstruction and classification objective, operating effectively under only 10\% visibility. This unified design enables both low-level recovery and high level understanding, outperforming single purpose baselines under low visibility conditions.

\section{Conclusion}
\label{sec:conclusion}

In this work, we introduced SMARC, a unified framework for Surface MAterial Reconstruction and Classification under extreme visual sparsity. Unlike traditional reconstruction or recognition pipelines that rely on dense visual input, SMARC learns to infer both texture and semantics from a single, continuous 10\% patch of the image. By coupling a Partial Convolutional U-Net with a lightweight classification head, the model jointly performs spatial inpainting and material recognition within one coherent architecture. Extensive experiments on the Touch and Go dataset demonstrate SMARC’s ability to achieve high fidelity reconstruction (17.55 dB PSNR) and robust material classification (85.10\%), outperforming leading vision transformers and autoencoder baselines. Future extensions of SMARC will focus on improving reconstruction quality, classification accuracy and computational efficiency. We plan to explore parameter pruning, mixed-precision training, and lightweight attention modules to reduce model complexity and inference time while preserving visual quality. 

\vspace{1cm} 
\noindent\textbf{\large Acknowledgement:}
The authors acknowledge the support and resources provided by the Bioinspired Robotics, AI, Imaging and Neurocognitive Systems (BRAINS) Laboratory and Predictive Analytics and Technology Integration (PATENT) Laboratory at The University of Alabama. 





\end{document}